%% file: main.tex
\documentclass[sigconf]{acmart}

\AtBeginDocument{%
  }

\copyrightyear{2023} 
\acmYear{2023} 
\setcopyright{acmlicensed}\acmConference[MM '23]{Proceedings of the 31st ACM International Conference on Multimedia}{October 29-November 3, 2023}{Ottawa, ON, Canada}
\acmBooktitle{Proceedings of the 31st ACM International Conference on Multimedia (MM '23), October 29-November 3, 2023, Ottawa, ON, Canada}
\acmPrice{15.00}
\acmDOI{10.1145/3581783.3612293}
\acmISBN{979-8-4007-0108-5/23/10}

\usepackage[ruled, vlined]{algorithm2e}
\usepackage{algpseudocode}

\usepackage{amsmath,bm}

\usepackage{wrapfig}
 \usepackage{mathtools}
\usepackage{multirow}
\usepackage{stfloats}
\usepackage{bbding}
\usepackage{threeparttable}
\usepackage{graphicx}
\usepackage[marginal]{footmisc}

\usepackage{multicol}
\usepackage{enumitem}
\usepackage{fontawesome}

\usepackage{color}
\usepackage{hyperref}
\hypersetup{
	colorlinks=true,
	linkcolor=red,
	filecolor=blue,      
	urlcolor=red,
	citecolor=green,
}

\usepackage{xcolor}

\acmSubmissionID{2627}

\begin{document}

\title{Progressive Spatio-temporal Perception for Audio-Visual Question Answering}


\author{Guangyao Li}
\orcid{0000-0002-2179-8555}
\affiliation{%
  \institution{Renmin Uniiversity of China,}
  \city{Beijing}
  \country{China}
}
\email{guangyaoli@ruc.edu.cn}

\author{Wenxuan Hou}
\orcid{0009-0009-0896-3729}
\affiliation{%
  \institution{Renmin Uniiversity of China,}
  \city{Beijing}
  \country{China}
}
\email{wxhou@ruc.edu.cn}

\author{Di Hu}
\orcid{0000-0002-7118-6733}
\authornotemark[1]
\affiliation{%
  \institution{Renmin Uniiversity of China,}
  \city{Beijing}
  \country{China}
}
\email{dihu@ruc.edu.cn}



\begin{abstract}
Audio-Visual Question Answering (AVQA) task aims to answer questions about different visual objects, sounds, and their associations in videos. 
Such naturally multi-modal videos are composed of rich and complex dynamic audio-visual components, where most of which could be unrelated to the given questions, or even play as interference in answering the content of interest. 
Oppositely, only focusing on the question-aware audio-visual content could get rid of influence, meanwhile enabling the model to answer more efficiently. In this paper, we propose a \textbf{P}rogressive \textbf{S}patio-\textbf{T}emporal \textbf{P}erception \textbf{Net}work (\textbf{PSTP-Net}), which contains three modules that progressively identify key spatio-temporal regions \emph{w.r.t.} questions. Specifically, a \textit{temporal segment selection module} is first introduced to select the most relevant audio-visual segments related to the given question. Then, a \textit{spatial region selection module} is utilized to choose the most relevant regions associated with the question from the selected temporal segments. To further refine the selection of features, an \textit{audio-guided visual attention module} is employed to perceive the association between auido and selected spatial regions. 
Finally, the spatio-temporal features from these modules are integrated for answering the question. 
Extensive experimental results on the public MUSIC-AVQA and AVQA datasets provide compelling evidence of the effectiveness and efficiency of \textbf{PSTP-Net}.
\vspace{-0.5em}
\end{abstract}

\vspace{-0.75em}


\vspace{-0.75em}




\maketitle

\begin{figure}[t]
     \centering
     \vspace{1.4em}
     \includegraphics[width=0.445\textwidth]{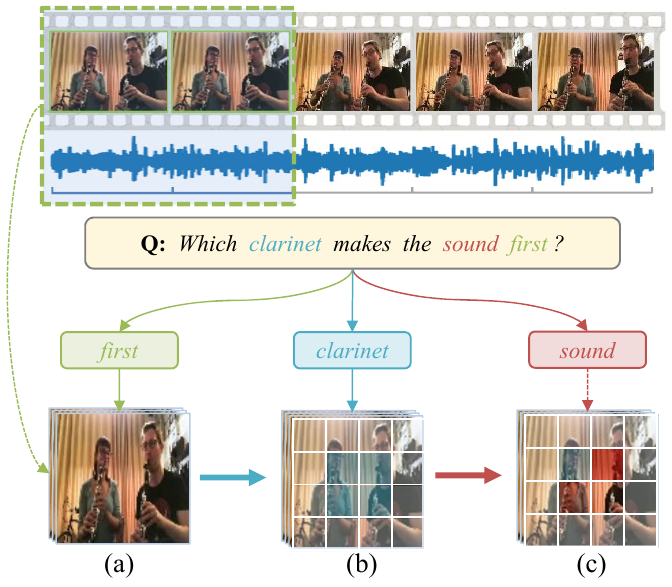}
     \vspace{-1.5em}
     \caption{Locating the relevant temporal audiovisual snippets and spatial sounding objects associated with a given question is critical for audiovisual scene understanding. 
     For instance, to achieve this when dealing with an input video, the process involves:
     (a) locating temporal segments relevant to the question;
     (b) identifying the spatial region most relevant to the question on selected segments;
     (c)  utilizing the selected audio features to perform sound-aware perception.
     }
     \label{fig:teaser}
     \vspace{-1.5em}
\end{figure}

\vspace{-0.3em}
\section{Introduction}
\vspace{-0.25em}
We are constantly surrounded by visual and auditory information in our daily life, and humans perceive the world by simultaneously processing and integrating visual and auditory inputs~\cite{holmes2005multisensory, wei2022learning}. As a widely used medium for recording and reflecting reality, natural scene videos typically convey important event information through visual and auditory streams. Audio-visual learning has therfore emerged as one of the key components in the multimedia community, garnering significant interest from researchers in the past decade~\cite{tian2018audio, alamri2019audio, tian2020unified, hu2021class, yun2021pano, li2022learning, yang2022avqa, hou2023towards, peng2022balanced, fan2020pyslowfast, jiang2022dhhn}.
\footnote{
\begin{itemize}[leftmargin=*]
\vspace{-0.75em}
\item Di Hu is the corresponding author, and also a Research Fellow of Metaverse Research Center, Renmin University of China, Beijing, P.R.China, 100872.
\item Code is available at \href{https://github.com/GeWu-Lab/PSTP-Net}{https://github.com/GeWu-Lab/PSTP-Net}.
\vspace{-0.75em}
\end{itemize} 
}

In particular, Audio-Visual Question Answering (AVQA)~\cite{yun2021pano, li2022learning, yang2022avqa} task involves answering questions related to intricate audio-visual scenes, which has gained much attention from researchers as a valuable and challenging task. Yun \emph{et al}.~\cite{yun2021pano} proposed the Pano-AVQA dataset, which comprises 360-degree videos and corresponding question-answer pairs. The Pano-AVQA dataset covers two types of question-answer pairs: spherical spatial relations and audio-visual relations, enabling the exploration of panoramic scene understanding. Li \emph{et al}.~\cite{li2022learning} introduced the MUSIC-AVQA dataset, a large-scale dataset that focuses on promoting spatio-temporal reasoning in dynamic and long-term audio-visual scenes, such as music performances. They first ground the sounding region visually, then perform spatio-temporal reasoning using attention mechanisms to perform effective question answering. To address the complexity of audio-visual relationships in real-life scenarios and the diverse range of audio-visual daily activities, Yang \emph{et al}.~\cite{yang2022avqa} proposed a large-scale AVQA dataset for multi-modal understanding of audio-visual objects and activities in realistic scenes in videos. These works provide benchmark platforms for evaluating complex audio-visual scene understanding and reasoning, and have achieved significant progress in advancing the task.

Despite the significant progress made in AVQA, there are still several challenges that need to be addressed. \textbf{Firstly}, the task involves audio-visual understanding over long videos, which suffer from heavy information redundancy and huge computational cost. Existing AVQA explorations typically use the uniform sampling strategy to reduce redundancy and computational cost, while may lose some valuable information. \textbf{Secondly}, localizing relevant regions of the question in the video is still challenging. Some VQA methods use pretrained object detection models to localize key objects, but since some special categories (\emph{e. g.}, \textit{souna}) in AVQA-related datasets are not included in datasets used for pretraining these models, they are still unable to effectively locate the relevant regions for the AVQA task. \textbf{Thirdly}, the lack of supervised information on spatial visual objects and sound poses a challenge for the model to associate visual targets with sound in the video, making it difficult to find potential sound-aware regions.
\textbf{Therefore}, identifying video segments relevant to the question, extracting relevant visual regions from these segments, and determining whether they produce sound are crucial for exploring the AVQA task. 
For instance, as shown in Fig.~\ref{fig:teaser}, when answering the audio-visual question "\textit{Which clarinet makes the sound first?}" for an instrumental ensemble, one needs to focus on the temporal segment related to "\textit{first}", locate object "\textit{clarinet}" and determine which clarinet produced the "\textit{sound}".

To address the above challenges, we propose a \textbf{P}rogressive \textbf{S}patio-\textbf{T}emporal \textbf{P}erception
\textbf{Net}work (PSTP-Net) to explore critical temporal segments and sound-aware regions among the complex audio-visual scenarios progressively.
Firstly, the content related to the question is usually scattered in partial segments of the video instead of the whole sequence. Hence, we propose a \textbf{T}emporal \textbf{S}egment \textbf{S}election \textbf{M}odule (TSSM)  that utilizes cross-modal attention to identify the most several relevant temporal segments for the given question. 
Secondly, identifying visual regions that are pertinent to the question within the selected key segments can aid in understanding spatial semantics. To achieve this, we propose the \textbf{S}patial \textbf{R}egion \textbf{S}election \textbf{M}odule (SRSM) to select the most relevant patches by using an attention-based ranking strategy. It enables a more effective comprehending of the spatial context presented in the video.
Thirdly, the sound and location of the visual source can reflect the spatial relationship between audio and visual modalities, which can help to learn audio-visual associations in complex scenarios. We propose an \textbf{A}udio-guided \textbf{V}isual \textbf{A}ttention \textbf{M}odule (AVAM) module to model such cross-modal associations of audio signals and patches selected by the SRSM module. 
Finally, we fuse the above selected audio and visual modalities and obtain a joint representation for question answering. Extensive experimental results demonstrate that our proposed PSTP-Net can achieve precise spatial-temporal perception and outperforms pervious methods with only a limited number of learnable parameters and FLOPs. Our proposed approach takes a step toward more effective and efficient audio-visual scene understanding. Our contributions can be summarized as follows:
\begin{itemize}[leftmargin=*]
\item The proposed PSTP-Net excels in perceiving key temporal segments, identifying visual regions relevant to the posed question, and cross-modal association between audio signal and visual patches, which facilitate the scene understanding over audio, visual, and text modalities.
\item The PSTP-Net comprises several modules that progressively perceive key spatio-temporal regions, effectively reducing redundant information and computational cost.
\item Extensive experiments and ablation studies on the MUSIC-AVQA and AVQA benchmarks demonstrate the effectiveness and efficieny of the proposed framework.
\end{itemize}

\section{Related Works}

\subsection{Audio visual scene understanding }
Audio-visual scene understanding tasks focus on exploring audio-visual joint perception~\cite{wei2022learning}. Several tasks fall under this domain, including sound source localization~\cite{senocak2018learning, hu2021class, hu2022self}, action recognition~\cite{gao2020listen}, event localization~\cite{tian2018audio, brousmiche2021multi}, video parsing~\cite{tian2020unified, wu2021exploring}, 
segmentation~\cite{zhou2022audio, liu2023annotation}, dialog~\cite{alamri2019audio, schwartz2019simple}, question answering~\cite{yun2021pano, li2022learning, yang2022avqa}, \textit{etc}. To achieve effective audio-visual scene understanding, it is essential to establish associations between audio and visual modalities. The sound source localization task~\cite{zhao2018sound, Gao_2021_CVPR, gan2020music, zhou2022sepfusion} and the audio-visual segmentation task~\cite{zhou2022audio} aim to locate visual objects relevant to sound, usually learning with annotations of sounding objects. However, such annotations are usually not provided in the AVQA task. To address this, researchers have experimented with constructing positive and negative pairs~\cite{li2022learning} or designing different multi-modal fusion models~\cite{yun2021pano, yang2022avqa} to enhance audio-visual correlation learning. Furthermore, some adapter-inserted transformer models have been proposed to enhance model representation capabilities.

However, despite these explorations and efforts, the fundamental problem of audio-visual correlation still remains unresolved. In this work, we propose to calculate the similarity between the audio input and its corresponding visual patch-level features. This enables the identification of the visual patches that most relevant to the sound, as well as the accurate visual sound source position localization.

\subsection{Question answering}
As a typical video understanding task, question answering explores fine-grained scene understanding, including Audio Question Answering (AQA)~\cite{fayek2020temporal}, Visual Question Answering (VQA)~\cite{antol2015vqa}, and Audio Visual Question Answering (AVQA)~\cite{li2022learning}.

The uni-modal question answering tasks represented by AQA~\cite{zhang2017speech, fayek2020temporal, li2023multi} and VQA~\cite{antol2015vqa, zhao2017video, kim2020modality, fan2019heterogeneous, yu2019activitynet,  Xiao_2021_CVPR, gao2023mist} have been extensively studied. However, they are restricted to a single modality of either sound or visual scenes, making it difficult to perceive and reason about natural audio-visual video content.
Yun \emph{et al}.~\cite{yun2021pano} proposed the Pano-AVQA dataset, which includes 360-degree videos and their corresponding question-answer pairs, aimed at exploring understanding of panoramic scenes.
Li \emph{et al}.~\cite{li2022learning} presented the MUSIC-AVQA dataset, which is a large-scale dataset with rich audio-visual components and interaction, designed to promote research on spatio-temporal reasoning in dynamic and long-term audio-visual scenes.
Considering that real-life scenarios contain more complex relationships between audio-visual objects and a greater variety of audio-visual daily activities, Yang \emph{et al}.~\cite{yang2022avqa} proposed a large-scale AVQA dataset to facilitate multimodal understanding of audio-visual objects and activities in real scenes captured in videos.

Although the AVQA task has garnered significant attention~\cite{LAVISH_CVPR2023, chen2023valor}, existing works are still in the preliminary exploration stage. Our work identifies temporal segments and localize spatial sound patches related to the given question progressively to explore complex audio-visual scene understanding.

\begin{figure*}[t]
     \centering
     \includegraphics[width=0.96\textwidth]{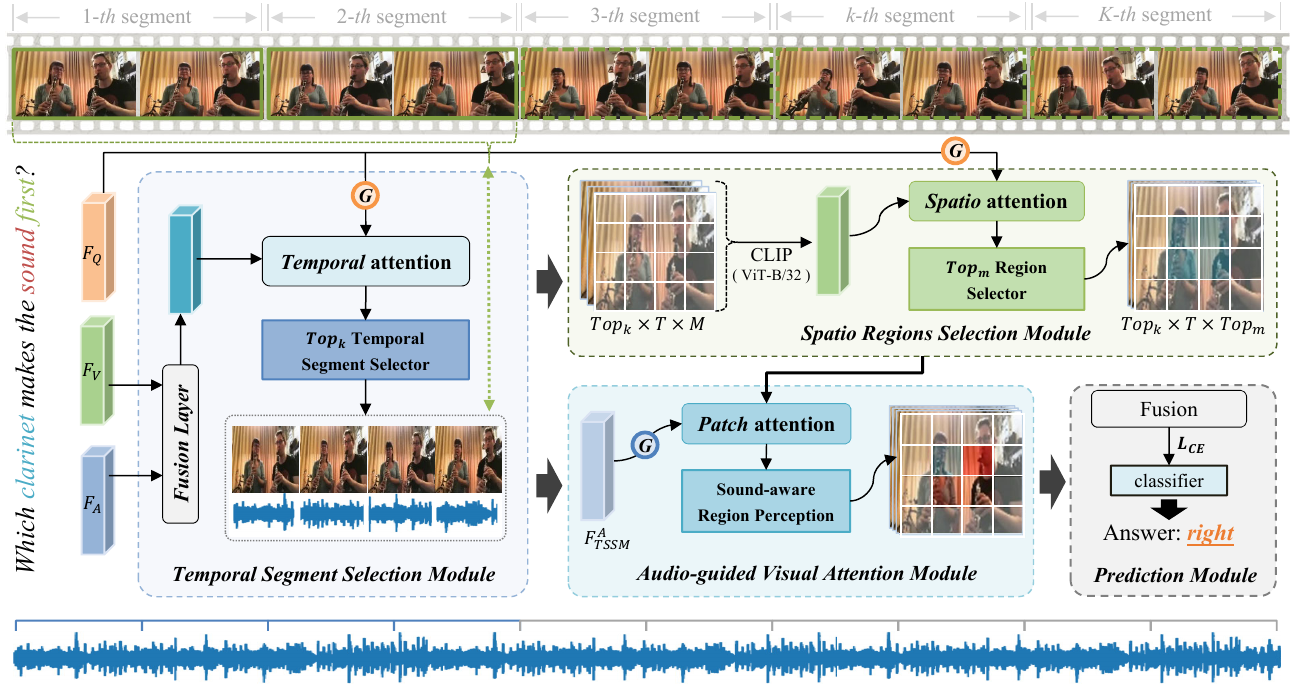}
     \vspace{-1.2em}
     \caption{
     Our proposed PSTP-Net model has a simple yet effective pipeline. 
     Firstly, the video is divided into $K$ segments, and we use a pre-trained model to extract audio, visual, and question features. 
     Then, we calculate the similarity between the temporal segment features and the input question feature to highlight the $Top_k$ relevant segments with the given question. 
     Next, we choose the $Top_m$ most relevant patches associated with the question from the selected segment features. 
     Afterwards, we perform audio-guided attention to perceive potential sound regions through the selected patches. 
     Finally, we aggregate the extracted audio, visual, and question features through multimodal fusion to predict the answer to the input question.
     (\includegraphics[scale=0.46]{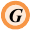}, \includegraphics[scale=0.46]{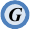} represent \textit{question-guided} and \textit{audio-guided}, respectively.)
     }
     \label{fig:pipe}
     \vspace{-0.9em}
\end{figure*}

\section{Method}
In this section, we will present our proposed \textbf{P}rogressive \textbf{S}patio-\textbf{T}emporal \textbf{P}erception Network (PSTP-Net), which achieves precise perception of relevant temporal segments and spatial sound sources, resulting in a better scene understanding than its unimodal counterparts utilizing audio or video. An overview of the proposed framework is illustrated in Fig.~\ref{fig:pipe}.

\vspace{-0.6em}
\subsection{Input Representation}
\vspace{-0.1em}
Given an input video sequence containing both visual and audio tracks, we first divide it into $K$ non-overlapping audio and visual segment pairs $\{V_k, A_k\}_{k=1}^K$, where each segment is $T$-seconds long. $V_k$ and $A_k$ denote visual and audio content in the same segment, respectively.
For the $k$-th audio-visual segment pair, it contains $\{v_t, a_t\}_{t=1}^T$ snippet, where each snippet is $\textit{1}$-second.
Subsequently, we partition each visual frame into $M$ patches and append a special $[CLS]$ token to the beginning of the first patch.
The question sentence $Q$ is tokenized into $N$ individual words $\{q_n\}_{n=1}^N$.
Note that we assume a frame sampling rate of $1fps$ for the the formula clarity.

\textbf{Audio Representation}.
For each audio snippet $a_t$, we use the pre-trained VGGish~\cite{gemmeke2017audio} model to extract the audio feature as $f_a^t \in \mathbb{R}^{D}$, where $D$ is the feature dimension. The pretrained VGGish model is a VGG-like 2D CNN network that trained on the AudioSet~\cite{gemmeke2017audio}, employing over transformed audio spectorgrams. 
The audio representation is extracted offine and the model is not fine-tuned. 
Then, the $k$-th audio segment features are extracted as $F_a^k=\{f_a^1, f_a^2, ..., f_a^T\}$, 
where $F_a^k \in \mathbb{R}^{T \times D}$.
Finally, the audio features can be fromulated as $F_A = \{F_a^1, F_a^2, ..., F_a^K\}$.

\textbf{Visual Representation}.
For each visual snippet $v_t$, we sample a fixed number of frames.
Then we apply pre-trained CLIP~\cite{radford2021learning}, with frozen parameters, extract both frame-level and patch-level features as $f_v^t$ and $f_p^t$ on video frames, respectively, where $f_v^t \in \mathbb{R}^{D}$, $f_p^t \in \mathbb{R}^{M \times D}$ and $M$ is patch numbers of one frame.
For the $k$-th segment, its \textbf{frame-level} features are extracted as $F_v^k=\{f_v^1, f_v^2, ..., f_v^T\}$, where $F_v^k \in \mathbb{R}^{T \times D}$,
and its \textbf{patch-level} features are extracted as $F_p^k=\{f_p^1, f_p^2, ..., f_p^T\}$, where $F_p^k \in \mathbb{R}^{T \times M \times D}$.
Finally, the visual frame-level and patch-level features can be fromulated as $F_V = \{F_v^1, F_v^2, ..., F_v^K\}$, $F_P = \{F_p^1, F_p^2, ..., F_p^K\}$, respectively.

\textbf{Question Representation}.
Given an asked question $Q=\{q_n\}_{n=1}^N$, we first represent each word $q_n$ to a fixed-length vector with word embeddings, and then fed into pre-trained CLIP\cite{radford2021learning} model to get question-level features $F_Q$, where $F_Q\in\mathbb{R}^{1\times D}$.
Note that the first token pooling is used for extracting question features.

\begin{figure*}[ht]
     \centering
     \includegraphics[width=0.94\textwidth]{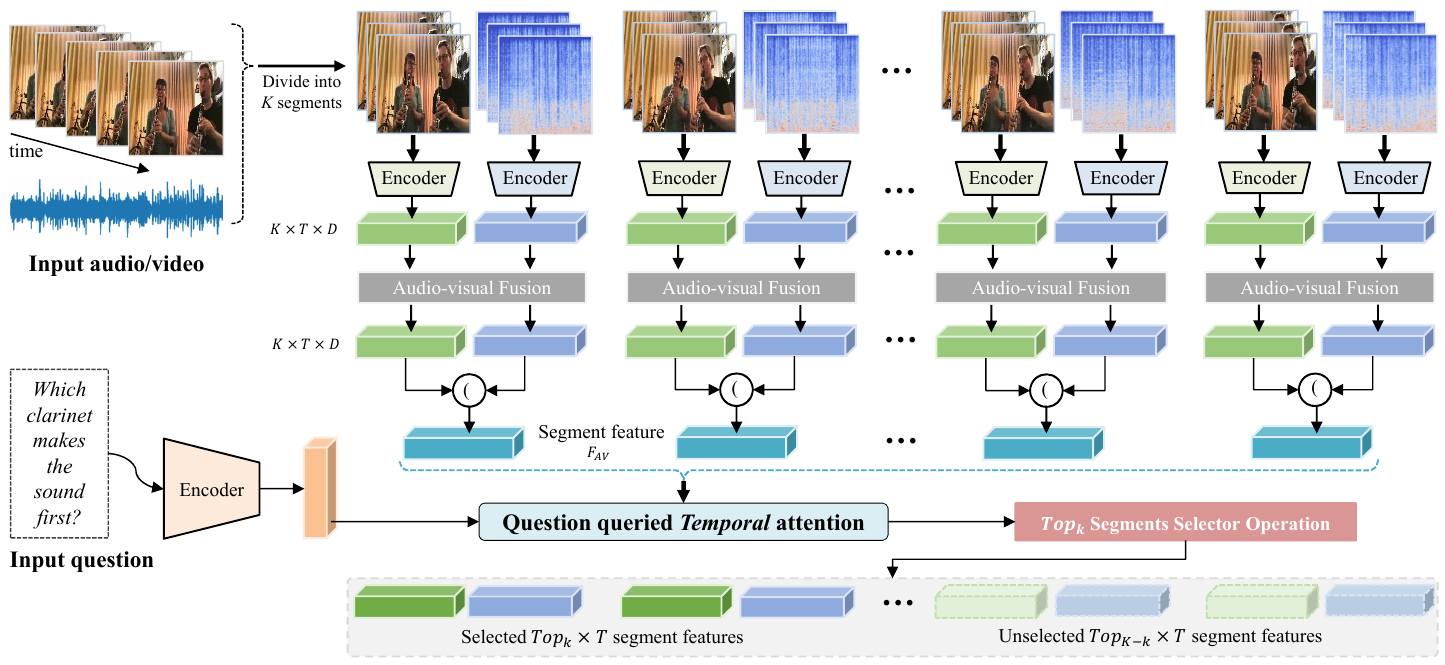}
     \vspace{-0.5em}
     \caption{The pipeline of the designed TSSM.     
     Firstly, TSSM performs self-attention and cross-modal perception on the audio and visual feature sequences in each segment. 
     Then, we perform pooling and concatenation operations on both modalities' features in the temporal dimension to obtain segment-level joint representation. Finally, we calculate the similarity between the question and the fused segments to find the ones most relevant to the question.
     }
     \label{fig:TSSM}
     \vspace{-0.75em}
\end{figure*}

\subsection{Temporal Segment Selection}
To hightlight the $Top_k$ key segments that are closely associated to the question, we propose a Temporal Segments Selection Module (TSSM), which is designed for attending critical temporal segments among all $K$ segments. An overview of the proposed TSSM is illustrated in Fig.~\ref{fig:TSSM}.
To simulataneously capture multimodal temporal contexts, we use an audio-visual fusion strategy, named $AVF$, which is designed for providing temporal feature interactions.
\textbf{For the $\boldsymbol{k}$-th segment}, an $AVF$ will be learned from audio and visual features $\{f_a^t, f_v^t\}_{t=1}^T$ to update $f_a^t$ and $f_v^t$, repectively. 
Firstly, the transformer encoder is employed to aggregate both within-modality and cross-modality information as: 
\begin{align}
\label{avf1}
\phi _{sa}(f_a^t, F_a^k, F_a^k) 
= Softmax (\frac{ f_a^t F_a^k}{\sqrt{d}}) F_a^k,
\\
\phi _{ca}(f_a^t, F_v^k, F_v^k)
= Softmax (\frac{ f_a^t F_v^k}{\sqrt{d}}) F_v^k,
\end{align}
where the scaling factor $d$ is equal to the audio feature dimension, $\phi _{sa}(\cdot)$ and $\phi _{ca}(\cdot)$ are self-attention and cross-modal attention functions, respectively.
Then we aggregate these representations to update audio and visual features $\hat{f}_a^t$, $\hat{f}_v^t$:
\begin{align}
\label{avf2}
\hat{f}_a^t
= f_a^t + \phi_{sa}(f_a^t, F_a^k, F_a^k) + \phi_{ca}(f_a^t, F_v^k, F_v^k),\\
\hat{f}_v^t
= f_v^t + \phi_{sa}(f_v^t, F_v^k, F_v^k) + \phi_{ca}(f_v^t, F_a^k, F_a^k),
\end{align}
where $F_a^k=\{f_a^1, f_a^2, ..., f_a^T\}$ and $F_v^k=\{f_v^1, f_v^2, ..., f_v^T\}$.
Afterward, the $k$-th audio and visual segment feature is updated to $\hat{F}_a^k$ and $\hat{F}_v^k$, where $\hat{F}_a^k = \{\hat{f}_a^1, \hat{f}_a^2, ..., \hat{f}_a^T\}$ and $\hat{F}_v^k = \{\hat{f}_v^1, \hat{f}_v^2, ..., \hat{f}_v^T\}$.
Then we perform a pooling operation on $\hat{F}_a^k, \hat{F}_v^k$ in the temporal dimension to obtain the $k$-th segment feature representation $\overline{F}_a^k, \overline{F}_v^k$, where $\overline{F}_a^k \in \mathbb{R}^{1 \times D}$, $ \overline{F}_v^k \in \mathbb{R}^{1 \times D}$.
Then, $\overline{F}_a^k$ and $\overline{F}_v^k$ are concatenated, and an $FC$ and activation layer are used to generate a joint representation, denoted as $\overline{F}{_{av}^k\in \mathbb{R}^{1 \times D}}$.
Finally, all segment-level features $\overline{F}{_{av}^k}$ are aggregated into $\overline{F}{_{AV}}=\{\overline{F}{_{av}^1}, \overline{F}{_{av}^2}, ..., \overline{F}{_{av}^K}\}$, where $\overline{F}{_{AV} \in \mathbb{R}^{K \times D}}$.

Giving the segment features $\overline{F}{_{AV}}$ and the input question feature $F_Q$, we first perform temporal cross-modal attention on $\overline{F}_{AV}$, and $F_Q$ after using a linear projection layer formulated as:
\vspace{-0.35em}
\begin{align}
\label{topkindex}
\tilde{F}_{AV}, W_{AV} 
= MultiHead(F_Q, \overline{F}{_{AV}}, \overline{F}{_{AV}}),
\end{align}
\vspace{-0.3em}
where $\tilde{F}_{AV}$ is updated from $\overline{F}_{AV}$, 
$W_{AV}$ is the average attention weights of all heads,
$W_{AV} \in (0, 1)^{K}$.
Then we conduct $Top_k$ feature selection over $K$ segments.
To be specific, we employ a selection operation, denoted as $\Psi_{TSSM}$, to pick out the segments with highest attenton weights and their index as follow:
\vspace{-0.35em}
\begin{align}
\label{selectk}
F_{TSSM}, \Omega_{TSSM}= \Psi_{TSSM}(\tilde{F}_{AV}, W_{AV}, Top_k),
\end{align} 
where $F_{TSSM}$ is selected temporal feature,  $F_{TSSM} \in \mathbb{R}^{Top_k\times T\times D}$, $\Omega_{TSSM}$ is the index position corresponding to the $Top_k$ highest weights in $W_{AV}$, 
$\Omega_{TSSM}\in \{0,1,..., k\text{-1}\}^{Top_k}$.

Furthermore, we select the visual patch-level features $F_{TSSM}^P \in \mathbb{R}^{Top_k \times M \times D}$ and audio features $F_{TSSM}^A \in \mathbb{R}^{Top_k \times D}$ from $F_P$ and $F_A$, respectively, according to $\Omega_{TSSM}$. 
For formula clarity, we record the index of the snippet corresponding to these selected segments and sort them from 1 to $\mathit{\Gamma}$, where $\mathit{\Gamma} = T \times Top_k$.
Then $F_{TSSM}^P$ and $F_{TSSM}^A$ are redescribed as:
\begin{align}
\label{topkattn}
F_{TSSM}^P = \{f_p^1, f_p^2, ..., f_p^{\mathit{\Gamma}}\}, 
F_{TSSM}^P\in \mathbb{R}^{\mathit{\Gamma} \times M \times D},
\end{align}
\vspace{-1.5em}
\begin{align}
\label{topkattn2}
F_{TSSM}^A = \{f_a^1, f_a^2, ..., f_a^\mathit{\Gamma}\}, 
F_{TSSM}^A\in \mathbb{R}^{\mathit{\Gamma} \times D}.
\end{align}
In next sections, we utilize the question feature to identify key spatial regions from selected patch-level features $F_{TSSM}^P$, and use selected audio features $F_{TSSM}^A$ to perform sound-aware perception.

\begin{figure}[t]
     \centering
     \includegraphics[width=0.47\textwidth]{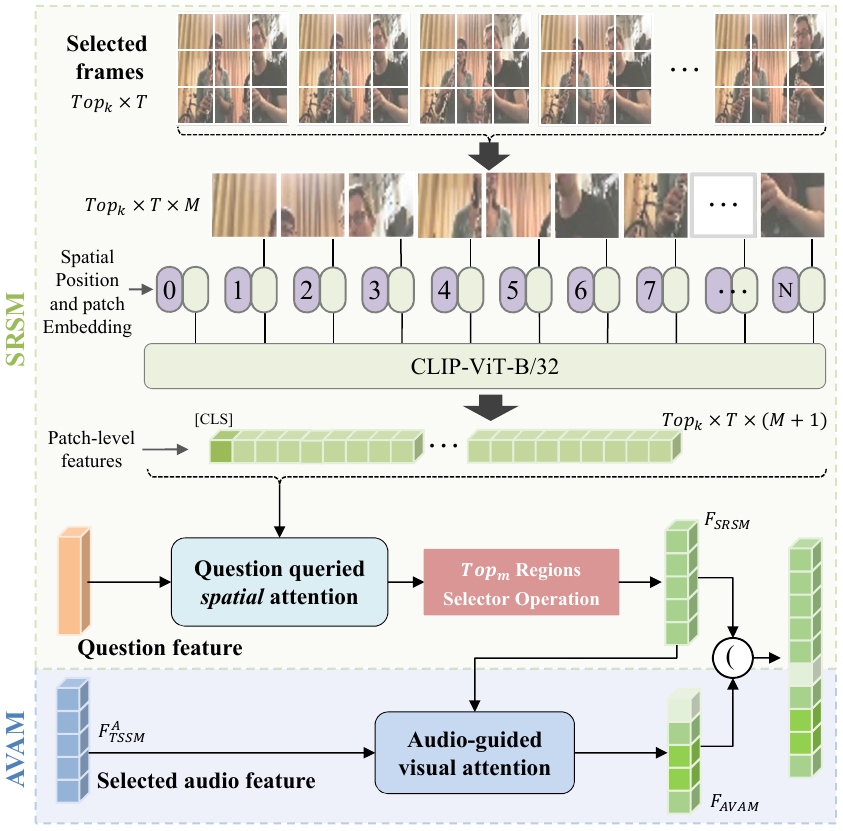}
     \vspace{-0.5em}
     \caption{Framework of SRSM (top part) and AVAM (bottom part). 
     For the selected temporal segments, the SRSM first locates the patches most relevant to the question on each frame. Then, the AVAM performs sound perception on the selected patches corresponding to the audio of each frame. 
     Finally, the features obtained from different modules are aggregated to a joint feature representation.}
     \label{fig:trsm}
\end{figure}

\vspace{-0.25em}
\subsection{Spatial Region Selection}
To identify visual regions that are pertinent to the question,
we design a Spatial Region Selection Module (SRSM) to choose the $Top_m$ most relevant regions associated with the question from the selected frame patch-level features, as shown in the top of Fig.~\ref{fig:trsm}.
\textbf{For the $\boldsymbol{\gamma}$-th frame}, its patch-level feature is $f_p^{\gamma} \in\mathbb{R}^{M\times D}$. We perform spatial cross-modal attention over $f_p^{\gamma}$ and $F_Q$ after using a linear projection layer, which can be formulated as:
\begin{align}
\label{topsindex}
\hat{f}{_p^{\gamma}},\ W_{\gamma} = MultiHead(F_Q, f_p^{\gamma}, f_p^{\gamma}),
\\
\tilde{f}_p^{\gamma}, \Omega_{\gamma} = \Psi_{SRSM}(\hat{f}{_p^{\gamma}}, W_{\gamma}, Top_m),
\end{align}
where $\hat{f}{_p^{\gamma}}$ is updated from $f_p^{\gamma}$, 
$W_{\gamma} \in (0,1)^M$ is the average attention weights of all heads,
$\Psi_{SRSM}$ adopts a selection strategy similar to $\Psi_{TSSM}$,
$\Omega_{\gamma} \in \{0,1,...,M\}^{Top_m}$ is the index position corresponding to the $Top_m$ highest weights in $W_{\gamma}$, and 
$\tilde{f}_p^{\gamma} \in \mathbb{R}^{Top_m\times D}$.
Then we stack all $\mathit{\Gamma}$ selected patch-level features to obtain the selected patch-level features of all selected frames $F_{SRSM}$:
\begin{align}
\label{topm}
F_{SRSM}=
\{\tilde{f}{_p^1}, \tilde{f}{_p^{2}}, ..., \tilde{f}{_p^{\mathit{\Gamma}}}\},
F_{SRSM}\in \mathbb{R}^{\mathit{\Gamma} \times Top_m\times D}.
\end{align}
\vspace{-1.5em}

\vspace{-0.25em}
\subsection{Audio-guided Visual Attention}
Considering that the sound and the location of its visual source usually reflects the patial association between audio and visual modality, we introduce the Audio-guided Visual Attention Module (AVAM) to decompose the complex scenarios into concrete audio-visual association, which performs attention-based patch-level sound source perception, as shown in the bottom of Fig.~\ref{fig:trsm}.
\textbf{For the $\boldsymbol{\gamma}$-th frame},  
giving its selected patches feature $f_p^{\gamma}$ along with its corresponding audio features $\tilde{f}_a^{\gamma}$, we perform cross-modal attention to perceive potential sound regions after using a linear projection layer, which can be formulated as:
\begin{align}
\label{topkindex2}
\tilde{f}{_{ap}^{\gamma}} = MultiHead(f_a^{\gamma}, \tilde{f}_p^{\gamma}, \tilde{f}_p^{\gamma}),
\end{align}
where $\tilde{f}{_{ap}^{\gamma}}$ is updated from $\tilde{f}_p^{\gamma}$, 
$\tilde{f}{_{ap}^{\gamma}} \in \mathbb{R}^{Top_m\times D}$.
Then we stack all $\mathit{\Gamma}$ sound-aware patch-level features to obtain:
\begin{align}
\label{topmaudio}
F_{AVAM} = \{\tilde{f}{_{ap}^1}, \tilde{f}{_{ap}^{2}}, ..., \tilde{f}{_{ap}^{\mathit{\Gamma}}}\},
F_{AVAM}\in \mathbb{R}^{\mathit{\Gamma} \times Top_m\times D}.
\end{align}
Then, the audio guided visual contextual embeddings are more capable of predicting correct answers.

Thus far, we have progressively identified the key temporal segments that are most relevant to the input question, as well as the spatial regions that are most relevant to the input question on the key segments, and its potential sound-aware patches.

\vspace{-0.25em}
\subsection{Lightweight Global Perception}
Considering that answering some questions requires incorporating information from the entire video, we propose a Lightweight Global Perception Module (LGPM) to address this issue. 
Specifically, for the input audio feature $F_A$ and visual frame-level feature $F_V$, we apply a cross-modal attention mechanism similar to $AVF$ over temporal dimension for both modalities, which can be formalized as:
\begin{align}
\label{LGPM}
F_{LGPM}^A
= F_A + \phi_{sa}(F_A, F_A, F_A) + \phi_{ca}(F_A, F_V, F_V),
\\
F_{LGPM}^V
= F_V + \phi_{sa}(F_V, F_V, F_V) + \phi_{ca}(F_V, F_A, F_A),
\end{align}
where $F_{LGPM}^A$, $F_{LGPM}^V$ are updated from $F_A, F_V$.

\subsection{Multimodal Fusion and Answer Prediction}
We concatenate the visual features $F_{TSSM}$, $F_{SRSM}$, $F_{AVAM}$, and $F_{LGPM}$ obtained from \textbf{TSSM}, \textbf{SRSM}, \textbf{AVAM}, and \textbf{LGPM}, respectively, to obtain viusal fusion feature $\overline{F}{_V}$.
Similarly, we concatenate the audio features $F_{SRSM}^A$ and $F_{LGPM}^A$ generated by the above modules to obtain audio fusion feature $\overline{F}{_A}$. 
Then, we concatenate $\overline{F}{_V}$ and $\overline{F}{_A}$ 
to obtain audio-visual feature embedding:
\begin{align}
\label{f1}
F_{aggregate} = ReLU(Concat[\overline{F}{_V}, \overline{F}{_A}]).
\end{align}
Then, we integrate audio-visual and question features by employing an element-wise multiplication operation to get the aggregated multimodal feature representation $\overline{F}_{aggregate}$. Concretely, we can formulate the fusion function as:
\begin{align}
\label{f2}
\overline{F}_{aggregate} = \boldsymbol{FC}(\delta(Pool(F_{aggregate}) \circ F_Q)),
\end{align}
where $\delta$ and $\boldsymbol{FC}$ represent $Tanh$ activation function and linear layer, respectively.
Finally, we utilize a linear layer and softmax function to output probabilities $p \in (0,1)^{C}$ for candidate answers. 
With the predicted probability vector and the corresponding ground-truth label, we can optimize our network by a cross-entropy loss.


\input{./tables/sota.tex}

\vspace{-0.6em}
\section{Experiments}

\vspace{-0.15em}
\subsection{Datasets}
The \textbf{MUSIC-AVQA} dataset~\cite{li2022learning} contains 9,288 videos covering 22 different instruments, with a total duration of over 150 hours and 45,867 Question-Answering (QA) pairs. Each video contains around 5 QA pairs in average. The questions are designed under multi-modal scenes containing 33 question templates covering 9 types. The MUSIC-AVQA dataset is well-suited for studying spatio-temporal reasoning for dynamic and long-term audio-visual scenes. 
The \textbf{AVQA} dataset~\cite{yang2022avqa} is designed for audio-visual question answering on general real-life scenario videos. It contains 57,015 videos from daily audio-visual activities, along with 57,335 QA pairs specially designed relying on clues from both modalities, where information from a single modality is insufficient or ambiguous. 
For both datasets, we adopt the official split of the two benchmarks into training, evaluation, and test sets.

\vspace{-0.5em}
\subsection{Implementation Details}
For the visual stream, we divide the video into $1$-second snippets and sample the corresponding frames at $1fps$, the pre-trained CLIP-ViT-B/32~\cite{radford2021learning} is used as the visual feature extractor to generate frame-level and patch-level 512-D feature vector for each visual snippet. For each $1$-second long audio snippet, we use a linear layer to process the extracted 128-D VGGish feature into a 512-D feature vector. For each question sentence, we extract its feature by the pre-trained CLIP-ViT-B/32~\cite{radford2021learning} and obtain a 512-D feature vector. In all experiments, we use the Adam optimizer with an initial learning rate of $1e-4$ and will drop by multiplying 0.1 every 10 epochs. We split all videos into 20 segments ($K=20$), in the proposed modules, we set $Top_k=7$, $Top_m=20$. Batch size and number of epochs are set to 64 and 30, respectively. We use the $thop$ library in PyTorch to calculate the model's parameters and FLOPs. Our model is trained on NVIDIA GeForce RTX 3090 and implemented in PyTorch.

\vspace{-0.75em}
\subsection{Quantitative Results}
\vspace{-0.25em}
To evaluate the effectiveness of the proposed PSTP-Net, we compare it with the existing method on the MUSIC-AVQA benckmark: 
AVSD~\cite{schwartz2019simple}, Pano-AVQA~\cite{yun2021pano}, AVST~\cite{li2022learning}, \emph{etc.} 
Tab~\ref{cmp} indicate that our method outperforms all comparison methods. Our method shows significant improvements in the subtask types of \textit{Audio-visual}, including \textit{Counting}, \textit{Localization}, \textit{Comparative}, and \textit{Temporal}. 
Specifically, compared to the AVST~\cite{li2022learning}, PSTP-Net achieves remarkable improvements of 3.35\%, 7.56\%, 7.12\%, and 3.18\% in the above-mentioned four complex question types, respectively. Additionally, the PSTP-Net shows a performance boost of 3.63\% and 2.09\% in the \textit{Counting} and \textit{Localization} subtasks of the visual modality, respectively, when compared to AVST~\cite{li2022learning}. 
The significant performance improvements indicate that our proposed PSTP-Net is effective in accurately identifying crucial temporal segments, spatial regions, and their corresponding sound-aware perception in videos.
We observe that the audio-only questions suffer from performance degradation, this could be because the visual content plays negative influence when answering audio-only questions. For example, the visual content tends to play as noise when answering the audio-only comparative-type question "\textit{Is the <Object1> louder than the <Object2>?}". Such phenomenon can be also found in some other methods\cite{alamri2019audio, yun2021pano}. This intriguing observation motivates us to explore strategies in future work that can achieve better performance on both single-modality and multi-modality aspects. 
In a word, the proposed PSTP-Net is capable of comprehending complex audio-visual scenes, especially in scenarios where combining audio and visual information is necessary to answer questions.

In Tab~\ref{flop2}, we present the efficiency of our proposed PSTP-Net model by evaluating it using two metrics, parameters and FLOPs, and comparing it with AVST~\cite{li2022learning}. 
Comparison results indicate that the proposed PSTP-Net achieves a significant reduction in training parameters (18.480M \emph{vs.} 4.297M) and FLOPs (3.188G \emph{vs.} 1.223G) compared to AVST, with a decrease of 76.73\%$\downarrow$ and 63.68\%$\downarrow$ in parameters and FLOPs, respectively. The main reason for this phenomenon is the \textit{spatial grounding module} designed for AVST, which aims to enhance audio-visual association by incorporating extra positive and negative sample pairs. However, this module heavily relies on $FC$ layers and spatial matrix calculations, resulting in a significant increase in the model's parameters and FLOPs. In contrast, the PSTP-Net effectively reduces the parameters and FLOPs by identifying critical temporal segments and spatial regions through its various designed modules, thus minimizing redundancy.
Notably, this reduction in computational requirements is accompanied by an improvement in accuracy compared to AVST, indicating the efficiency of our model. 
As illustrated in the Tab~\ref{flop2}, the PSTP-Net achieves better performance with lower parameters and FLOPs, showing its superior efficiency.
In summary, the PSTP-Net offers significant improvements over existing approaches and has the potential to advance the field of audio-visual scene understanding.

\input{./tables/FLOPsParams.tex}

\input{./tables/module.tex}

\subsection{Ablation Study}
In this subsection, we investigate how different configuration choices of our model affect the performance on the MUSIC-AVQA dataset.

\textbf{Effect of components in PSTP-Net}. We ablate key modules in PSTP-Net,  \emph{i.e.},\ TSSM, SRSM, AVAM, and LGPM, denoted as:
\begin{itemize}[leftmargin=*]
\item \textbf{PSTP-Net w/o. all}. 
All modules are removed except for the fusion of the input audio, video, and question features to verify whether the improved performance of PSPT-Net is due to the replacement of the ResNet-18 pre-trained model with CLIP-ViT-B/32.
Table ~\ref{module} shows that there is little performance difference when using two different backbones in the model \textit{PSTP-Net w/o. all}. However, compared to PSTP-Net, its performance drops to 69.47\% (4.05\%$\downarrow$), indicating that the performance improvement is not solely attributed to the replacement of ResNet-18 with CLIP-ViT-B/32 for feature extraction.
It is worth noting that we extract input features offline (\emph{i.e.}, extract input features before training) and ensure their output dimension is the same as ResNet-18, the parameters and FLOPs of the PSTP-Net remain unchanged.

\item \textbf{PSTP-Net w/o. TSSM}.
The temporal segment selection module is removed, which enables the model to locate key regions and develops sound-aware perception in all video frames. This results in significant temporal redundancy, leading to a decrease in model performance and an increase in FLOPs.  As shown in Tab~\ref{module}, the performance of the PSTP-Net drop to 72.98\%, with FLOPs increased by 60\% compared to PSTP-Net.

\item \textbf{PSTP-Net w/o. SRSM}.
The spatial region selection module is removed, causing the model to rely solely on the audio feature selected by TSSM to identify potential sound-aware areas in its corresponding video frames, which limits its ability to prioritize spatial areas relevant to the question. 
As shown in Tab~\ref{module}, compared with PSTP-Net, the performance of this model decreased to 72.92\%, indicating the importance of the spatial region selection module in improving performance.

\begin{figure}[t]
     \centering
     \includegraphics[width=0.45\textwidth]{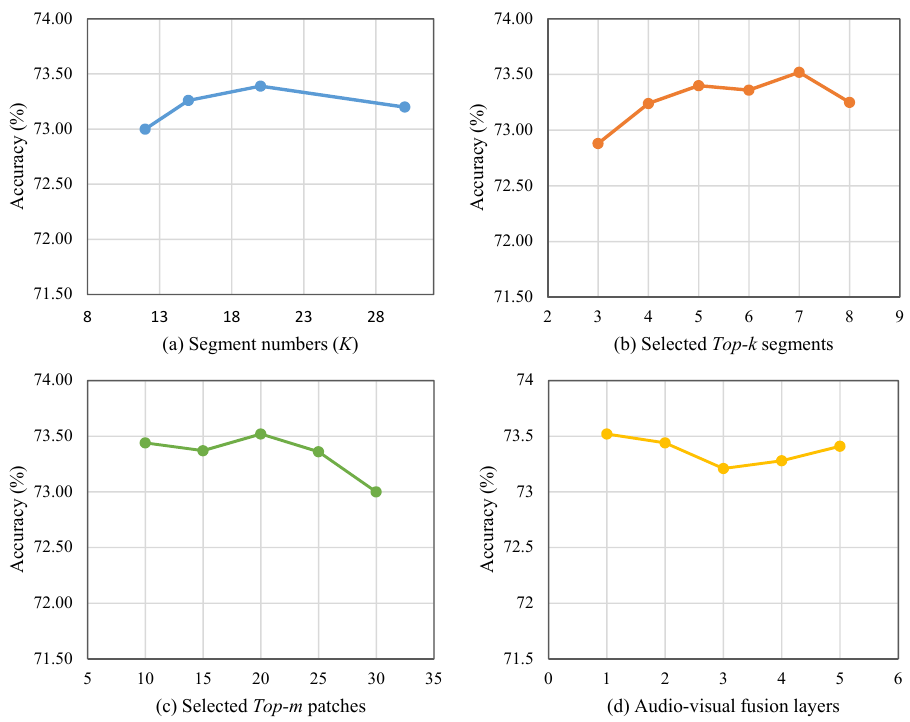}
     \vspace{-0.5em}
     \caption{Effects of different PSTP-Net configurations.}
     \label{fig:eff}
     \vspace{-1.5em}
\end{figure}

\item \textbf{PSTP-Net w/o. AVAM}.
The AVAM aims to perceive sound sources within the selected spatio-temporal regions and learn the association between audio and visual modalities. It is crucial for answering questions that require the perception of audiovisual correlations in the spatial context. For example, the accuracy of answering audio-visual localization questions is improved by 6.37\% when the AVAM is performed (71.80\% vs. 65.43\%). Additionally, the AVAM is a lightweight module, it only incur a small additional computational cost (10.3\% FLOPs of the PSTP-Net).

\item \textbf{PSTP-Net w/o. LGPM}.
The lightweight global perception module is removed, 
results in the model only attending to selected temporal segments and their corresponding regions. This caused the model to ignore global cues and makes it difficult to answer certain questions that require access to the entire video.
As shown in Tab~\ref{module}, compared to the model $w/o.$ LGPM, the performance of PSTP-Net improves by 0.85\%, with a low extra cost.

\begin{figure*}[t]
     \centering
     \includegraphics[width=0.92\textwidth]{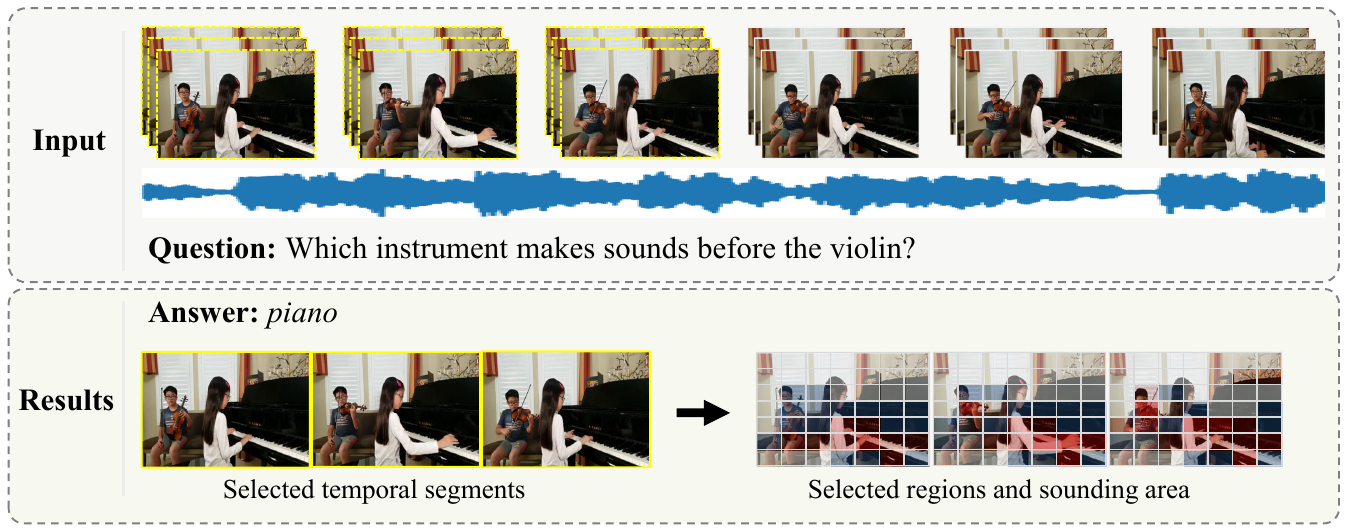}
     \vspace{-1em}
     \caption{Visualization results of the PSTP-Net. Based on the selection results of our method, the question-related area, the sounding area, and key timestamps are highlighted in the spatial and temporal dimensions, respectively, which indicates that our method can effectively model the spatio-temporal association across different modalities. }
     \label{fig:vis}
     \vspace{-0.75em}
\end{figure*}

\end{itemize}
Overall, each module contributes to better performance. When all modules are present, the proposed PSTP-Net achieves the best performance on the MUSIC-AVQA dataset.

\textbf{Effects of different PSTP-Net configurations}.
We also explore the impact of different configurations on model performance, including the number of temporal segments $K$, selected segments $Top_k$, selected patches $Top_m$, and audio-visual fusion layers.
We conducted all experiments on the MUSIC-AVQA and selected the best configuration on the validation set.
As indicated in Fig~\ref{fig:eff}(a), the model's performance first increases and then decreases with increasing $K$. This phenomenon can be attributed to the fact that events in long videos occur for a considerable duration. 
If the segment duration is too short, an event may be divided into fragments, making it incomplete and difficult to capture semantic information. Meanwhile, if the segment duration is too long, it will result in too much redundant noise accompanying the event.
Fig~\ref{fig:eff}(b) shows that the performance initially improves and then declines with the increase of Topk, displaying slight fluctuations. 
This could be because few segments may overlook essential question-related clues, while too many may introduce irrelevant redundant information. 
Fig~\ref{fig:eff}(c) displays the impact of the selected number of patches on the model. As the number of patches increases, the performance decreases, suggesting that the relevant target region occupies only a portion of space. 
Moreover, we analyzed the impact of the number of audio-visual fusion layers on the performance. 
As shown in Fig~\ref{fig:eff}(d), more transformer layers cannot achieve better performance while leads to an increase in FLOPs. We set the number of layers to 1 in the experiment.
In general, different configurations may cause slight performance fluctuations, but most results maintain above 73.00\%, demonstrating the stability of the proposed PSPT-Net.

\input{./tables/avqa.tex}

\vspace{-0.25em}
\subsection{Experiments on AVQA dataset}
\vspace{-0.15em}
To achieve more promising results, we conduct experiments with the proposed PSPT-Net on AVQA dataset. 
Specifically, we split all videos into 5 segments ($K=5$), and set select $Top_3$ temporal segments and $Top_25$ spatio regions most relevant to the question, respectively.
Then we aggregate these updated features with the features obtained from different modules to obtain a joint representation for answering questions. 
Results are present in Tab~\ref{avqa2}, the model achieves the best results compared to the ensemble $HCRN+HAVF$. 
Obviously, the proposed PSPT-Net has significant potential for improving the performance on AVQA dataset.

\vspace{-0.25em}
\subsection{Visualization Results}
\vspace{-0.15em}
We visualize some cases from the MUSIC-AVQA dataset in Fig~\ref{fig:vis}. Visualization results show that the PSTP-Net can effectively select video temporal segments, frame-level regions, and their corresponding sounding-aware patches that are relevant to the questions.

\vspace{-0.25em}
\section{Conclusion and Discussion}
\vspace{-0.15em}
In this work, we propose a novel Progressive Spatio-Temporal Perception Network (PSTP-Net) for the audio-visual question answering task. The PSTP-Net first utilizes a temporal segment selection module to identify key temporal segments relevant to the given question. Then, a spatial region selection module is designed to select regions most relevant to the question. Next, an audio-guided visual attention module is introduced to enable sound-aware perception on the selected regions. Additionally, a lightweight global perception module is performed to capture global information. Finally, we fuse features extracted from the above modules to obtain a joint representation for answering questions. Extensive evaluations on the MUSIC-AVQA and AVQA datasets demonstrate the effectiveness and efficiency of the proposed PSTP-Net.

\textbf{Discussion}.
The proposed PSTP-Net is distinct from AVST~\cite{li2022learning} and VALOR~\cite{chen2023valor}, as they rely on constructing additional positive-negative pairs or using extra-large audio-visual datasets to enhance sound source visual localization accuracy.
In contrast, we adopt a progressive strategy on a limited dataset to identify the most relevant temporal segments and corresponding key regions to the question, followed by sound-aware perception on these key regions. Our approach enables us to achieve good performance at a lower cost. 
Additionally, we use pre-trained models to extract features offline, eliminating the need to train model parameters during the training process, unlike LAVISH~\cite{LAVISH_CVPR2023}, which significantly reduces the demand for computational resources. 
Meanwhile, we notice that fine-tuning pre-trained transformer-based models (\emph{e.g.}, SwinV2-L~\cite{liu2022swin}) can improve model performance~\cite{LAVISH_CVPR2023}. This suggests that fine-tuning large pre-trained models are beneficial for downstream AVQA tasks, but their finetuning strategy has not been sufficiently explored so far.
Moreover, Large Language Models (LLMs) have shown their great potential in many tasks~\cite{wu2023visual}, especially in demonstrating strong reasoning abilities in QA tasks.
These challenges provide a wide scope for further exploration on AVQA task.

\vspace{-0.5em}
\section{Acknowledgments}
\vspace{-0.15em}
This research was supported by National Natural Science Foundation of China (NO.62106272), the Young Elite Scientists Sponsorship Program by CAST (2021QNRC001), in part by the Research Funds of Renmin University of China (NO. 21XNLG17)  and Public Computing Cloud, Renmin University of China.


\bibliographystyle{ACM-Reference-Format}
\bibliography{mm_refs}


\end{document}

%% file: tables/sota.tex
\begin{table*}[ht]
\begin{center}

\scalebox{0.975}{

\begin{tabular}{c|ccc|ccc|cccccc|c}
\hline
\multirow{2}{*}{\textbf{Method}} & \multicolumn{3}{c|}{\textbf{Audio}}  & \multicolumn{3}{c|}{\textbf{Visual}}  & \multicolumn{6}{c|}{\textbf{Audio-Visual}}  & \multirow{2}{*}{\textbf{Avg}} \\
  & \textbf{Count} & \textbf{Comp}  & \textbf{Avg}   & \textbf{Count} & \textbf{Local} & \textbf{Avg}   & \textbf{Exist} & \textbf{Count} & \textbf{Local} & \textbf{Comp}  & \textbf{Temp}  & \textbf{Avg}   &  \\ \hline
FCNLSTM~\cite{fayek2020temporal}  & 70.80  & 65.66  & 68.90  & 64.58  & 48.08  & 56.23  & 82.29  & 59.92  & 46.20  & 62.94  & 47.45  & 60.42  & 60.81  \\
GRU~\cite{antol2015vqa}  & 71.29  & 63.13  & 68.28  & 66.08  & 68.08  & 67.09  & 80.67  & 61.03  & 51.74  & 62.85  & 57.79  & 63.03  & 65.03  \\
HCAttn~\cite{lu2016hierarchical}  & 70.80  & 54.71  & 64.87  & 63.49  & 67.10  & 65.32  & 79.48  & 59.84  & 48.80  & 56.31  & 56.33  & 60.32  & 62.45  \\
MCAN~\cite{yu2019deep}  & \textbf{78.07} & 57.74  & 70.58  & 71.76  & 71.76  & 71.76  & 80.77  & 65.22  & 54.57  & 56.77  & 46.84  & 61.52  & 65.83  \\
PSAC~\cite{li2019beyond}  & 75.02  & 66.84  & 72.00  & 68.00  & 70.78  & 69.41  & 79.76  & 61.66  & 55.22  & 61.13  & 59.85  & 63.60  & 66.62  \\
HME~\cite{fan2019heterogeneous}  & 73.65  & 63.74  & 69.89  & 67.42  & 70.20  & 68.83  & 80.87  & 63.64  & 54.89  & 63.03  & 60.58  & 64.78  & 66.75  \\
HCRN~\cite{le2020hierarchical}  & 71.29  & 50.67  & 63.69  & 65.33  & 64.98  & 65.15  & 54.15  & 53.28  & 41.74  & 51.04  & 46.72  & 49.82  & 56.34  \\
AVSD~\cite{schwartz2019simple}  & 72.47  & 62.46  & 68.78  & 66.00  & 74.53  & 70.31  & 80.77  & 64.03  & 57.93  & 62.85  & 61.07  & 65.44  & 67.32  \\
Pano-AVQA~\cite{yun2021pano}  & 75.71  & \underline{65.99}  & \underline{72.13}  & 70.51  & \underline{75.76}  & 73.16  & \underline{82.09}  & 65.38  & 61.30  & 63.67  & 62.04  & 66.97  & 69.53  \\
AVST~\cite{li2022learning}  & \underline{77.78}  & \textbf{67.17} & \textbf{73.87} & \underline{73.52}  & 75.27  & \underline{74.40} & \textbf{82.49} & \underline{69.88}  & \underline{64.24}  & \underline{64.67}  & \underline{65.82}  & \underline{69.53}  & \underline{71.59}  
\\ \hline
\textbf{PSTP-Net}  & 73.97  & 65.59  & 70.91  & \textbf{77.15}  & \textbf{77.36} & \textbf{77.26} & 76.18  & \textbf{73.23}  & \textbf{71.80} & \textbf{71.79} & \textbf{69.00} & \textbf{72.57} & \textbf{73.52}  
\\ \hline
\end{tabular}

}

\caption{Results of different methods on the test set of MUSIC-AVQA. The top-2 results are highlighted.}
\vspace{-2.75em}
\label{cmp}
\end{center}
\end{table*}

%% file: tables/FLOPsParams.tex
\begin{table}[t]
\begin{center}

\scalebox{1}{

\begin{tabular}{c|cc|c}
\hline
\multirow{2}{*}{\textbf{Method}} & 
\multirow{2}{*}{\textbf{\begin{tabular}[c]{@{}c@{}}Training Param (M)↓\end{tabular}}} &
\multirow{2}{*}{\textbf{\begin{tabular}[c]{@{}c@{}}FLOPs (G)↓\end{tabular}}} &
\multirow{2}{*}{\textbf{Avg}} \\

    &    &    &    \\ 

\hline
AVST~\cite{li2022learning}    & 18.480    & 3.188     & 71.59    
\\
PSTP-Net    & 4.297    & 1.223     & 73.52    \\ 
\hline
\end{tabular}

}
\caption{Parameter and FLOPs of PSTP-Net and AVST.}

\label{flop2}
\end{center}
\vspace{-2.75em}
\end{table}

%% file: tables/module.tex
\begin{table}[t]
\begin{center}

\scalebox{0.85}{

\begin{tabular}{c|c|cc|c}
\hline
\multirow{2}{*}{\textbf{Method}} & \multirow{2}{*}{\textbf{\begin{tabular}[c]{@{}c@{}}Pre-trained\\ model\end{tabular}}} & \multirow{2}{*}{\textbf{\begin{tabular}[c]{@{}c@{}}Training\\ Params (M)↓\end{tabular}}} & \multirow{2}{*}{\textbf{\begin{tabular}[c]{@{}c@{}}FLOPs\\ (G)↓\end{tabular}}} & \multirow{2}{*}{\textbf{Avg}} \\
    &    &    &    &    \\ \hline
PSTP-Net w/o. all    & ResNet-18    & 1.403    & 0.144    & \multicolumn{1}{l}{69.01}    \\
PSTP-Net w/o. all    & CLIP-ViT-B/32    & 1.403    & 0.144    & 69.47    \\
PSTP-Net w/o. TSSM    & CLIP-ViT-B/32    & 3.250    & 1.945    & 72.98    \\
PSTP-Net w/o. SRSM    & CLIP-ViT-B/32    & 3.508    & 0.955    & 72.92    \\
PSTP-Net w/o. AVAM    & CLIP-ViT-B/32    & 3.770    & 1.212    & 73.26    \\
PSTP-Net w/o. LGPM    & CLIP-ViT-B/32    & 3.769    & 1.096    & 72.67    \\ \hline
PSTP-Net    & CLIP-ViT-B/32    & 4.297    & 1.223    & 73.52    \\ \hline
\end{tabular}

}

\caption{Different module results on MUSIC-AVQA test set. }

\label{module}
\end{center}
\vspace{-2.5em}
\end{table}

%% file: tables/avqa.tex
\begin{table}[t]
\begin{center}

\scalebox{0.9}{

\begin{tabular}{c|c|c}
\hline
\textbf{Method}    & \textbf{Ensemble} & \textbf{Total Accuracy (\%)} \\ \hline
HME~\cite{fan2019heterogeneous}       & HAVF~\cite{yang2022avqa}      & 85.0     \\
PSAC~\cite{li2019beyond}      & HAVF~\cite{yang2022avqa}      & 87.4     \\
LADNet\cite{li2019learnable}    & HAVF~\cite{yang2022avqa}      & 84.1     \\
ACRTransformer~\cite{zhang2020action} & HAVF~\cite{yang2022avqa}     & 87.8    \\
HGA~\cite{jiang2020reasoning}        & HAVF~\cite{yang2022avqa}      & 87.7     \\
HCRN~\cite{le2020hierarchical}      & HAVF~\cite{yang2022avqa}      & 89.0     \\
\textbf{PSTP-Net}       & --       & 90.2     \\ \hline
\end{tabular}

}

\caption{Results of different methods on the test of AVQA.}
\vspace{-3em}

\label{avqa2}
\end{center}
\end{table}